\documentclass[journal]{IEEEtran}

\usepackage{cite}

\ifCLASSINFOpdf
   \usepackage[pdftex]{graphicx}
\else
   \usepackage[dvips]{graphicx}
\fi

\usepackage{amsmath}
\usepackage{amssymb}
\usepackage{array}

\ifCLASSOPTIONcompsoc
  \usepackage[caption=true,font=normalsize,labelfont=sf,textfont=sf]{subfig}
\else
  \usepackage[caption=true,font=footnotesize]{subfig}
\fi

\usepackage{stfloats}
\usepackage{url}
\usepackage{booktabs}
\usepackage{multirow}
\usepackage{microtype}
\usepackage[breaklinks=true,colorlinks,bookmarks=false]{hyperref}
\usepackage[group-separator={,}, group-minimum-digits=4]{siunitx}

\begin{document}

\title{Learn to Interpret Atari Agents}

\author{Zhao~Yang,
        Song~Bai,
        Li~Zhang,
        and~Philip~H.S.~Torr\\\vspace*{8pt} 
        Technical Report$^\dagger$\\Spring 2019%
        
\thanks{$^\dagger$ Uploaded for archival purposes only. Compared to the previous version, this one contains some additional experiments.}%
}

\maketitle

\begin{abstract}
Deep reinforcement learning (DeepRL) agents surpass human-level performance in many tasks. However, the direct mapping from states to actions makes it hard to interpret the rationale behind the decision-making of the agents. In contrast to previous \emph{a-posteriori} methods for visualizing DeepRL policies, in this work, we propose to equip the DeepRL model with an innate visualization ability. Our proposed agent, named region-sensitive Rainbow (RS-Rainbow), is an end-to-end trainable network based on the original Rainbow, a powerful deep Q-network agent. It learns important regions in the input domain via an attention module. At inference time, after each forward pass, we can visualize regions that are most important to decision-making by backpropagating gradients from the attention module to the input frames. The incorporation of our proposed module not only improves model interpretability, but leads to performance improvement. Extensive experiments on games from the Atari 2600 suite demonstrate the effectiveness of RS-Rainbow.
\end{abstract}

\section{Introduction} \label{sec:introduction}
Understanding deep neural networks (DNNs) has been a long-standing goal of the machine learning community.
In the past, many efforts have been devoted to producing human-interpretable visual explanations~\cite{Simonyan2014Deep,deconv,guided-backprop,deeplift,fong,yarin,shen_SalAware_VOS,shen_SalDetFCN,meng_tnnls_sal} for CNN-based classification models (\emph{e.g.}, \cite{alexnet,simonyan2014very}).
With the advent of deep reinforcement learning (DeepRL)~\cite{atari13,dqn15}, there is now an increasing interest in developing methods for understanding DeepRL models.
Combining deep learning techniques and reinforcement learning algorithms, DeepRL leverages the strong representation capacity and approximation power of DNNs for optimizing the objective of return estimation or policy estimation~\cite{sutton}.
In modern applications where a state is represented by high-dimensional data (\emph{e.g.}, Atari 2600~\cite{atari-py}, VizDoom~\cite{vizdoom}, MuJoCo~\cite{MuJoCo}, \emph{etc}.), DeepRL needs to learn low-dimensional representations for the states as well as a policy that maximizes the long-term return.

As DeepRL does not optimize for a classification objective, previous interpretation methods developed for classification models are not readily applicable.
The approximation of the optimal state value or action distribution not only operates in a black-box manner, but encodes temporal information and environment dynamics.
The black-box and sequential nature of DeepRL models makes them inherently hard to understand.
In recent years, many efforts have been devoted to interpreting these complex models.
Most of the existing interpretation methods~\cite{dqn15,duel,icml16,icml18} are a-posteriori, aiming to explain a model after it has been trained.
For instance, some t-SNE-based methods~\cite{dqn15,icml16} employ game-specific human intuitions and expert knowledge in RL, while some other vision-based methods~\cite{duel} adopt a saliency approach.
The representative work of~\cite{icml18} takes a data-driven approach and illustrates policy responses to a fixed input masking function, which requires hundreds of forward passes per frame.
A common limitation to a-posteriori methods is that the deduced knowledge cannot be used to improve training.

In this work, we focus our study on the highly impactful family of deep Q-learning networks (DQN).
DQN was the first algorithm that achieved human-level control in a variety of Atari games, and it has been constantly innovated since, representing the state of the art in the discrete control domain today.
The most recent DQN variant, Rainbow~\cite{rainbow}, incorporates several complementary improvements on the original DQN and represents the state-of-the-art method for tackling the control problem in Atari games, among methods that do not exploit large-scale distributed training.

\begin{figure*}[!t]
  \centering
  \subfloat[]{
    \begin{minipage}[tb]{0.48\textwidth}
    \includegraphics[width=1\textwidth]{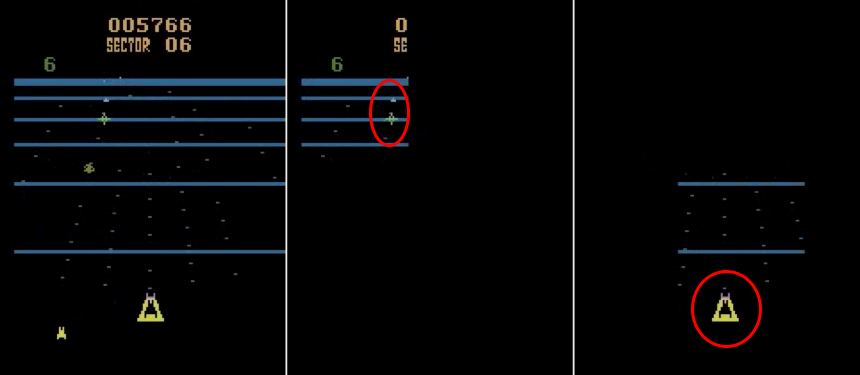}
    \end{minipage}
    \label{fig:1-1}
  }
  \subfloat[]{
    \begin{minipage}[tb]{0.48\textwidth}
    \includegraphics[width=1\textwidth]{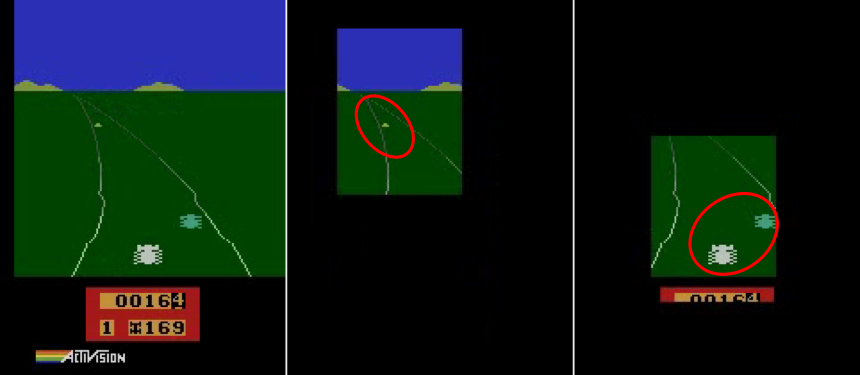}
    \end{minipage}
    \label{fig:1-2}
  }
  
  \subfloat[]{
    \begin{minipage}[tb]{0.48\textwidth}
    \includegraphics[width=1\textwidth]{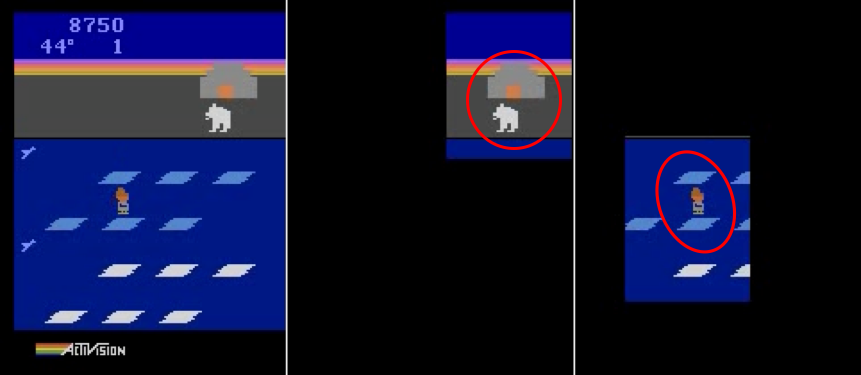}
    \end{minipage}
    \label{fig:1-3}
  }
  \subfloat[]{
    \begin{minipage}[tb]{0.48\textwidth}
    \includegraphics[width=1\textwidth]{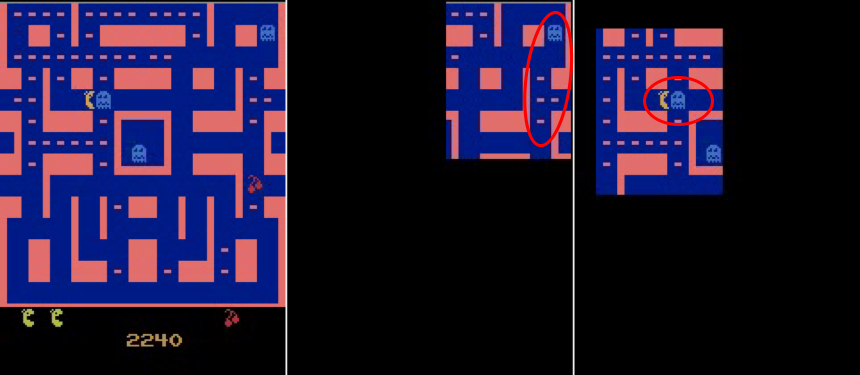}
    \end{minipage}
    \label{fig:1-4}
  }
  
  \subfloat[]{
    \begin{minipage}[tb]{0.48\textwidth}
    \includegraphics[width=1\textwidth]{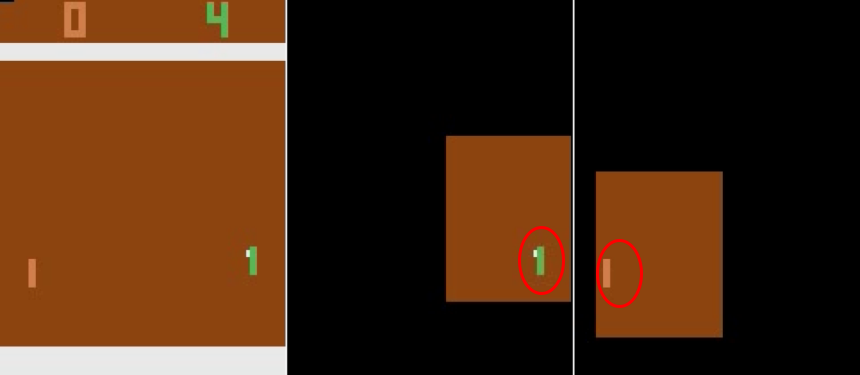}
    \end{minipage}
    \label{fig:1-5}
  }
  \subfloat[]{
    \begin{minipage}[tb]{0.48\textwidth}
    \includegraphics[width=1\textwidth]{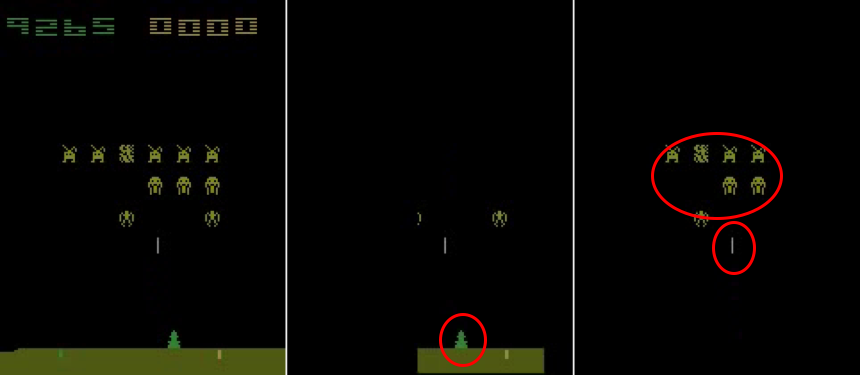}
    \end{minipage}
    \label{fig:1-6}
  }
  \caption{Visualizing Atari games. (a) Beam Rider. (b) Enduro. (c) Frostbite. (d) Ms. Pac-Man. (e) Pong. (f) Space Invaders. For each sub-figure, the left frame is the original frame, and each of the middle and the right frames shows a \emph{gaze} (defined in Section~\ref{sec:analysis}) of RS-Rainbow, obtained using gradient-based saliency. The agent learns multiple salient regions containing functional objects in the games, which are annotated with red circles.}
  \label{fig:1}
\end{figure*}

Specifically, we approach interpretation from a learning perspective and propose region-sensitive Rainbow (RS-Rainbow), to improve both the interpretability and performance of a DeepRL model.
To this end, RS-Rainbow leverages a region-sensitive module to estimate the importance of different regions on the screen, which is used to guide policy learning in an end-to-end fashion.
During the forward pass, multiple attention patterns are predicted, each of which assigns importance scores to regions according to their usefulness to action prediction.
Each attention pattern may favor different regions on the screen, and may be interpreted as representing the considerations of the agent when it is making a decision.
To produce visualizations, we backpropagate gradients from the most important region from each attention pattern to the input frames, arriving at several gradient-based saliency maps~\cite{Simonyan2014Deep} on the screen, which may provide useful information that helps us understand the agent's decision.
Figure~\ref{fig:1} illustrates such examples of visualizations (at one instance of time during the game) where the number of attention patterns is two.

In addition to being a learning-based approach for interpreting DeepRL models that requires no extra supervision, our method has the following advantages:
\begin{itemize}
  \item In contrast to previous methods~\cite{icml16,icml18}, RS-Rainbow can reflect what is actually being `looked' at in the agent's forward pass, without any human interventions.
  \item Besides supporting interpretation, quantitative evaluation of the RS-Rainbow on the Atari 2600 benchmark demonstrates that the adoption of the region-sensitive module can boost model performance. In comparison, previous a-posteriori methods are unable to do that.
  \item The region-sensitive module, the core component of RS-Rainbow, is a simple and efficient plug-in. It has the potential to be applied to other DQN-based models.
\end{itemize}

The rest of the paper is organized as follows.
In Section~\ref{background}, we provide a brief overview of the background; in Section~\ref{sec:approach}, we discuss the details of the proposed RS-Rainbow; in Section~\ref{sec:analysis}, we demonstrate the interpretability of RS-Rainbow; in Section~\ref{sec:experiments}, we quantitatively evaluate RS-Rainbow on several Atari games and Gym Retro; and in Section~\ref{sec:conclusion},
we summarize our main findings and discuss potential future directions.

\section{Background} \label{background}
\subsection{DQN and Rainbow}
Reinforcement learning (RL) concerns the problem of an agent learning to act in an environment with the goal of maximizing its return.
The environment is represented by states (finite or infinite), denoted as $S$, a transition rule from one state to the next (stochastic or deterministic), denoted as $P$, a reward function, denoted as $R$, and a set of actions available to the agent, denoted as $A$.
In any given state $s_t$ of the environment (partially or wholly observable), the agent chooses an action $a_t$ to act based on its current policy $\pi$; the action causes the environment to transition to a new state $s_{t+1}$, and the agent receives a scalar reward, represented as $r_{t+1}$.
The goal of the agent is to learn a policy that allows it to accumulate as much return as possible before the terminal state is reached.
The above model of the environment, the agent, and their interaction can be formally defined as a Markov decision process, denoted as $(S,A,P,R,\gamma)$, where $\gamma$ is a reward discounting factor~\cite{sutton}.

A value-based RL algorithm, called Q-learning, can be used to tackle the policy learning problem.
The objective of the Q-learning algorithm is to learn an action-value function $Q(s_t,a_t)$ that estimates the expected return of executing action $a_t$ in the state $s_t$ while following the current policy.
Given the optimal estimations of Q values of all action-state pairs, the optimal policy of the agent is then to take the action associated with the highest Q value at each state.
The action-value function $Q(s_t,a_t)$ conforms to the Bellman equation,
$$
Q(s_t,a_t) = \mathbb{E}[r_{t+1}+\gamma \mathop{\textrm{max}}_{a'}Q(s_{t+1},a')|s_t,a_t].
$$

If the number of states is tractable, then $Q(s_t,a_t)$ can be represented by a table, and Q values can be updated by temporal difference as follows, $Q(s_t,a_t) \leftarrow Q(s_t,a_t) + \alpha[r_{t+1}+\gamma \textrm{max}_{a'}Q(s_{t+1},a') - Q(s_t,a_t)]$, where $\alpha$ is a hyperparameter.
This learning strategy has a convergence guarantee~\cite{multi-step,td}.
In most real-world applications (such as the Atari games) however, the number of states is intractable.
In this case, we may want to use a function approximator to estimate the action-value function $Q(s_t,a_t)$.
DQN~\cite{atari13,dqn15} employs a deep convolutional neural network as the function approximator, called the \emph{online network}, and the parameters of this network are optimized by minimizing the following mean-squared error loss,
$$
L(\theta_i) = [r_{t+1}+\gamma \mathop{\textrm{max}}_{a'}Q(s_{t+1},a';\theta^-_i)-Q(s_t,a_t;\theta_i)]^2.
$$
Here, $(s_t,a_t,r_{t+1},s_{t+1})$ forms a training example and is uniformly sampled from an \emph{experience replay bank}, which stores a fixed amount of transitions of the game; the \emph{target network} parameters $\theta^-_i$ are an ``old'' copy of the online network parameters $\theta_i$ at any time: They are updated from the online network periodically and kept fixed during each period~\cite{dqn15}.
The above formulation implements the idea of the temporal difference updating method by optimizing a mean-squared error objective, and this is made possible by the introduction of the target network and the experience replay mechanism, which are also important to stabilizing training~\cite{dqn15}.
In this way, DQN exploits the power of deep neural networks to adapt Q-learning to work for much more complicated tasks.

Rainbow~\cite{rainbow} incorporates many extensions from the original DQN, each of which enhances a different aspect of the original model.
The extensions include double DQN~\cite{double-dqn}, dueling DQN~\cite{duel}, priority experience replay~\cite{per}, multi-step learning~\cite{multi-step}, distributional RL~\cite{distributional}, and noisy nets~\cite{noisy}.
Double DQN addresses the problem of over-estimation of the Q value in the original DQN by changing the target function from $r_{t+1} + \gamma \mathop{\textrm{max}}_{a'}Q(s_{t+1},a';\theta^-_i)$ to $r_{t+1} + \gamma Q(s_{t+1}, \mathop{\textrm{argmax}}_{a'}Q(s_{t+1},a';\theta_i);\theta^-_i)$, in which the selection and evaluation of an action are performed by two different networks.
Dueling DQN decomposes the estimation of Q into estimations of two separate variables, the state value and the action advantage, and decouples their estimations with two separate downstream branches.
Priority experience replay prioritizes the sampling of training data that have higher estimated learning returns.
Multi-step learning looks multiple steps ahead by replacing one-step rewards and states with their multi-step counterparts.
Noisy net introduces a novel exploration strategy that achieves better efficiency than $\epsilon$-greedy exploration.
It injects adaptable noises into the policy layers by incorporating extra random variables in the linear transformation, leading to state-dependent exploration.
The model jointly optimizes for reducing the interference of noises and maximizing the expected return, leading to the co-adaptation of exploration and exploitation.
In practice, as the policy converges, the model learns to ignore the injected noises.
In distributional RL, the distribution of the return replaces the expectation of the return as the immediate optimization objective.
Thus, rather than directly outputting the final expected return, Q, the model learns a distribution of the return over a fixed support set of discrete values.
Then Q is predicted by computing the expectation of the return from the learned distribution.
The adopted Kullback-Leibler divergence loss has a weak convergence guarantee as the return distributions conform to a Bellman equation.

\subsection{Understanding DeepRL}
Traditionally, interpreting RL systems may involve language generation via first-order logic~\cite{Dodson, Elizalde, Khan, Hayes}.
These approaches require the state space to be small and high-level state variables with interpretable semantics.
As such, they are not applicable to most modern DeepRL applications, such as vision-based Atari 2600 tasks~\cite{atari-py}.

In the context of DeepRL,~\cite{dqn15} and~\cite{icml16} propose to interpret DQN policies in the t-SNE~\cite{tsne} embedding space.
\cite{icml16} proposes the semi-aggregated Markov decision process (SAMDP) method, which visualizes hierarchical spatio-temporal abstractions learned in the policy embeddings of the agent.
By extracting embeddings from the last policy layer and projecting them into t-SNE~\cite{tsne} space, SAMDP can discover meaningful clusters that can be described by game-specific attributes.
Furthermore, by tracing state transitions across thousands of states, SAMDP discovers frequent paths taken by the agent across the aforementioned clusters, from which one can summarize the high-level skills of the agent.
The manual selection of game-specific attributes makes SAMDP dependent on human knowledge and intuition for good performance.
In addition, extracting these attributes from simple emulators like Atari is particularly challenging without appropriate interface support.
While high-level abstractions are informative to RL experts, a user without relevant theoretical backgrounds may find them hard to understand.

The work in~\cite{icml18} utilizes perturbation-based saliency~\cite{deeplift} to visualize pixel importance in an asynchronous advantage actor-critic (A3C) model~\cite{a3c16}.
It applies a masking function at fixed dense locations on the input frame and observes the impact on the target output, measured by the Euclidean distance.
Such methods can be computationally inefficient as each perturbation requires a separate forward pass through the network. Therefore, hundreds of forward passes are required for computing saliency for a single frame.
Some work~\cite{deeplift} points out that saliency~\cite{Simonyan2014Deep,guided-backprop} tends to underestimate feature importance.

Sorokin~\emph{et al}.~\cite{darq} incorporate the hard and soft attention mechanisms~\cite{show-attend-tell} used for image caption generation into deep recurrent Q-network~\cite{hausknecht2015deep}, in which the policy layers in the original DQN are replaced with a recurrent neural network.
Their method computes a set of attention weights over spatial locations in feature maps and globally aggregates features into a single vector for representing the state, and it is shown to improve performance on two of the five tested Atari games.
The attention weights can be visualized on game frames.
In RS-Rainbow, we implement a region-sensitive module that preserves the spatial information in feature maps and is capable of focusing on multiple regions, and we further develop a saliency-based localization method that is able to identify the agent's area of interest in the input frames, thus producing visualizations that help us interpret the decision-making rationale of the agent.

\section{Proposed Approach} \label{sec:approach}
\begin{figure*}[!t]
  \centering
  \includegraphics[width=0.9\linewidth]{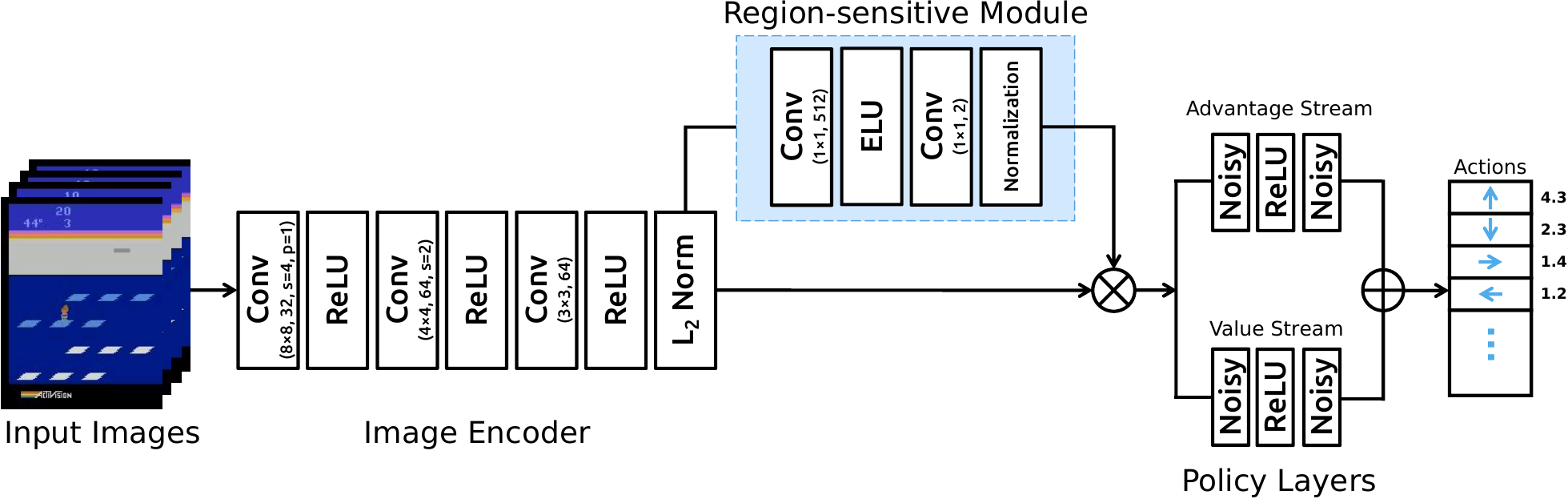}
  \caption{The architecture of the proposed RS-Rainbow. At each training step, four consecutive game frames are sampled and input to the model. The image encoder consists of three convolutional layers each followed by ReLU~\cite{relu} nonlinearity except for the last layer, which is followed by $L_2$ normalization. The region-sensitive module consists of two convolutional layers, where ELU~\cite{elu} nonlinearity follows the first layer and normalization (\emph{e.g.}, softmax) follows the second layer. The outputs of the image encoder and the region-sensitive module are combined via Hadamard product and fed to the advantage stream and the value stream to produce the final action scores.}
  \label{fig:model}
\end{figure*}
In this section, we describe our motivation in Section~\ref{sec:motivation}, then introduce the architecture of RS-Rainbow in Section~\ref{sec:architecture}, and finally demonstrate its capability of producing interpretable visualizations in Section~\ref{sec:visualization}.

\subsection{Motivation} \label{sec:motivation}
We outline three main considerations that motivate the design of RS-Rainbow.
First, in principle, pixels on the screen do not all contain useful information for value prediction.
For example, functional objects are critical while the background is less relevant.
Second, the relevance of an object depends on the specific state.
For instance, an unimportant object in the background may become important in some states when it is associated with reward signals, which can happen due to environment determinism.
Third, humans tend to play a game by looking at specific regions with high strategic values rather than considering all information on the entire screen.

Thus, we are interested in the following questions.
Will exploiting the relevance of objects in an environment benefit policy learning in DeepRL, for instance, by enabling the learning of a more useful state representation?
If so, how can we learn such relevance information without access to additional supervision signals?
And once learned, can object relevance help us better understand the inference process of a DeepRL agent?
In the following sections, we describe our approach to answering those questions.

\subsection{Architecture} \label{sec:architecture}
The complete architecture of RS-Rainbow is illustrated in Figure~\ref{fig:model}, which consists of an image encoder, the region-sensitive module, and policy layers with a value stream and an advantage stream.

As in Rainbow~\cite{rainbow}, our image encoder is a three-layer CNN with ReLU nonlinearities~\cite{relu}.
At each time step $t$, a stack of four consecutive frames $\textbf{S}\in\mathbb{R}^{4\times 84\times 84}$ is drawn from the replay memory.
The image encoder takes $\textbf{S}$ as input, and outputs the image embedding $\textbf{I}\in\mathbb{R}^{64\times 7\times 7}$, where $64$ denotes the number of channels and the two $7$s denote the height and the width, respectively.
We apply $L_2$ normalization to $\textbf{I}$ along the channel dimension to ensure scale invariance.

In the region-sensitive module, we employ two layers of $1\times1$ convolutions with ELU non-linearity~\cite{elu}.
The region-sensitive module takes $\textbf{I}$ as input, and outputs score maps $\textbf{A}=[\textbf{A}_1,\textbf{A}_2,...,\textbf{A}_N]\in\mathbb{R}^{N\times 7\times 7}$, where $N$ is the number of score maps, each of which has size $7\times 7$.
Each element on a score map corresponds to a spatial location on $\textbf{I}$, and describes the importance of features at that location.
Then, score maps $\textbf{A}$ are passed to a normalization layer to be turned into probability distributions.
In our experiments, we implement the normalization layer using either the softmax function or the sigmoid function.
The final probability distributions after normalizing $\textbf{A}$ are denoted as $\textbf{P}=[\textbf{P}_1,\textbf{P}_2,...,\textbf{P}_N]$, where $\textbf{P}_n\in\mathbb{R}^{1\times 7\times 7}$ is the $n$-th ($1\leq n\leq N$) probability distribution.

Each $\textbf{P}_n$ highlights a unique view of the agent about the importance of the regions.
$\textbf{P}_n$ can be viewed as a pattern detector that learns to respond to a pattern in the game by assigning high importance scores to regions.
At the same time, the most important area according to $\textbf{P}_n$ contains the most salient visual features relevant to decision-making.
After end-to-end training, $\textbf{P}$ learns several distributions of importance of regions, which complement each other.
Aside from the game rewards, no other supervision is needed for learning $\textbf{P}$.

From each $\textbf{P}_n$, we produce an image embedding, $\textbf{F}_n$, which is a unique representation of the state.
$\textbf{F}_n$ is obtained by multiplying $\textbf{P}_n$ and $\textbf{I}$ element-wise (broadcasting along the channel dimension), which can be denoted as $\textbf{F}_n=\textbf{P}_n\odot \textbf{I}.$
Hence, $\textbf{F}_n$ is of the same shape as $\textbf{I}$.
To obtain the final state representation, we add up all $\textbf{F}_n$s as $\textbf{F}=\sum_{n=1}^N \textbf{F}_n.$
In summary, from the original image embedding $\textbf{I}$, we estimate $N$ sets of importance scores each of which is a distribution of the image regions, and using which, we scale $I$ at each spatial location, then aggregate the $N$ scaled image representations to form the final representation of the state.

The region-sensitive module is related to the broader concept of \emph{attention}~\cite{jointly-align, attention-all-you-need}, which first gained popularity in methods addressing the neural machine translation task, and then quickly expanded to many subjects in computer vision and machine learning~\cite{ocnet,non-local,attention_lstm_gesture_rec_tnnls,meng_graph_reg,attention-receptive-fields-tnnls,stack,show-attend-tell}.
Different from the more popular formulation of attention as a mapping function from a query and a key-value pair to an aggregated output, our region-sensitive module seeks to estimate the regional importance at each location and simply uses that importance estimation to scale the original image features, which improves policy learning in vision-based DeepRL tasks, as we show in the experiments.

Finally, the policy layers consist of an advantage stream and a value stream, the outputs of which are combined to estimate the state-action value Q.
Each stream is implemented as two noisy linear layers~\cite{noisy} with ReLU~\cite{relu} non-linearity.
Finally, Q values are calculated as the mean of a learned distribution over a fixed support set of discrete return values, from which the policy can be derived.
\begin{figure}[!t]
  \centering
  \subfloat[]{
    \begin{minipage}{0.15\textwidth}
    \includegraphics[width=1\textwidth]{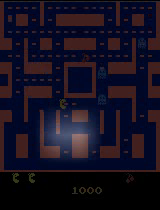}
    \end{minipage}
    \label{fig:vis_grid}
  }
  \subfloat[]{
    \begin{minipage}{0.15\textwidth}
    \includegraphics[width=1\textwidth]{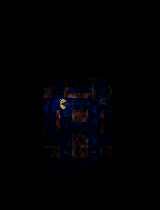}
    \end{minipage}
    \label{fig:vis_raw}
  }
  \subfloat[]{
    \begin{minipage}{0.15\textwidth}
    \includegraphics[width=1\textwidth]{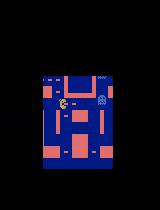}
    \end{minipage}
    \label{fig:vis_bin}
  }
  \caption{Three ways to produce visualizations with RS-Rainbow. (a) Importance score overlay. (b) Soft saliency mask. (c) Binary saliency mask.}
  \label{fig:vis_compare}
\end{figure}

\subsection{Visualization} \label{sec:visualization}
Based on the region-sensitive module, we can highlight salient regions and use them to interpret why the agent has made a decision.
In the following (and Figure~\ref{fig:vis_compare}), we describe three ways of producing visualizations with RS-Rainbow.

The first (see Figure~\ref{fig:vis_grid}) is to upsample the probability map, $\textbf{P}_n$, to the input resolution and overlay it on the stack of input frames (we use the last frame).
Since $\textbf{P}_n$ has spatial size $7\times7$, this approach effectively treats the screen as a $7\times7$ grid and assumes that the receptive field of each element in $\textbf{P}_n$ corresponds to a grid cell, which is actually not the case.
Therefore, the problem with this approach is that the localization is highly inaccurate.

In the second and the third alternative approaches, we mask the screen with a \emph{soft} and a \emph{binary} saliency map, respectively.
The saliency map is computed by taking the derivative of the largest score in $\textbf{A}_n$ with respect to the input frames (similarly, we use the last frame).
This type of saliency is called gradient-based saliency~\cite{Simonyan2014Deep}, and can be expressed mathematically as follows, $\textbf{G}_n=\frac{\partial \max_l(\textbf{A}_{nl})}{\partial \textbf{S}}, \textbf{G}_n\in\mathbb{R}^{4\times84\times84}$, where $l$ indexes spatial locations in $\textbf{A}_{n}\in \textbf{A}$.
Since gradients $\textbf{G}_n$ are unbounded in range and direction, we take the absolute values of $\textbf{G}_n$ and normalize the absolute values to the range of $[0, 1]$.
The final results are the saliency scores.
An interesting thing to note here is that while $\textbf{G}_n$ corresponds to the size of the screen, only pixels lying in the receptive field of the chosen element (the largest) from $\textbf{A}_{n}$ has non-zero values---the gradients are zeros everywhere else.
The saliency scores can be used as a soft mask (Figure~\ref{fig:vis_raw}), and if we turn each non-zero score into $1$ (thresholding at $0.0$), then they can be turned into a hard mask assuming 0/1 values everywhere (Figure~\ref{fig:vis_bin}).

To produce visualizations, we multiply the soft saliency mask and the binary saliency mask with the last input frame, respectively.
As shown in Figure~\ref{fig:vis_raw} and Figure~\ref{fig:vis_bin}, while both masks can locate salient objects and there are no essential differences between them, the binary saliency mask is more friendly to visualization (compare the two visualizations to see), and thus we use the hard binary mask in our analyses on the Atari agents in the next section.
Note that these attention/saliency patterns are automatically learned, which is different from existing a-posteriori methods.
Interested readers can refer to ~\cite{icml16,icml18} for more details.

\section{Atari Analysis} \label{sec:analysis}
\begin{figure*}[!ht]
  \centering
  \subfloat[General]{
    \begin{minipage}[tb]{0.32\textwidth}
    \includegraphics[width=1\textwidth]{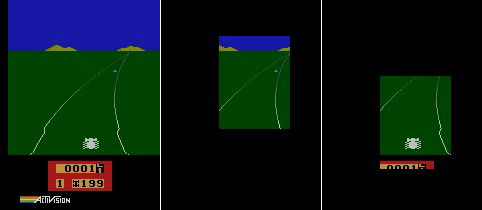}
    \end{minipage}
    \label{fig:en-1}
  }
  \subfloat[General]{
    \begin{minipage}[tb]{0.32\textwidth}
    \includegraphics[width=1\textwidth]{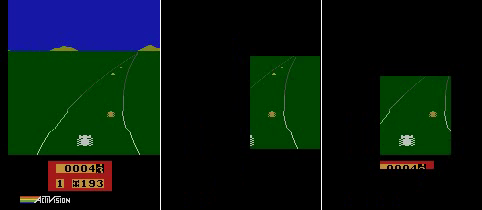}
    \end{minipage}
    \label{fig:en-2}
  }
  \subfloat[General]{
    \begin{minipage}[tb]{0.32\textwidth}
    \includegraphics[width=1\textwidth]{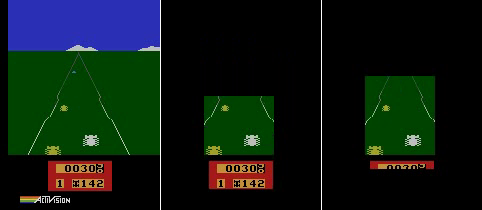}
    \end{minipage}
    \label{fig:en-3}
  }
  
  \subfloat[Counting down]{
    \begin{minipage}[tb]{0.32\textwidth}
    \includegraphics[width=1\textwidth]{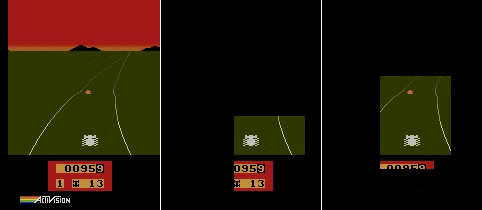}
    \end{minipage}
    \label{fig:en-4}
  }
  \subfloat[Counting down]{
    \begin{minipage}[tb]{0.32\textwidth}
    \includegraphics[width=1\textwidth]{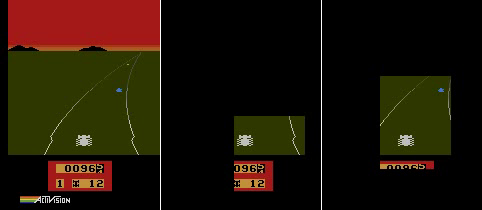}
    \end{minipage}
    \label{fig:en-5}
  }
  \subfloat[Counting down]{
    \begin{minipage}[tb]{0.32\textwidth}
    \includegraphics[width=1\textwidth]{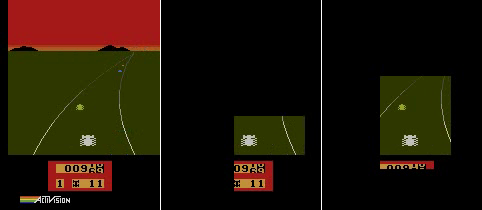}
    \end{minipage}
    \label{fig:en-6}
  }
  
  \subfloat[Slacking]{
    \begin{minipage}[tb]{0.32\textwidth}
    \includegraphics[width=1\textwidth]{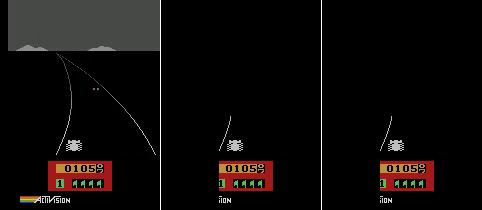}
    \end{minipage}
    \label{fig:en-7}
  }
  \subfloat[Prepping]{
    \begin{minipage}[tb]{0.32\textwidth}
    \includegraphics[width=1\textwidth]{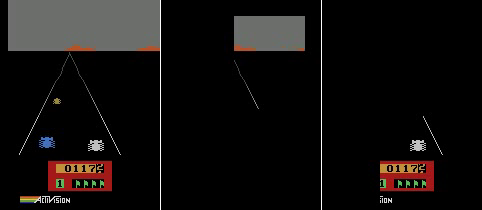}
    \end{minipage}
    \label{fig:en-8}
  }
  \subfloat[Smog]{
    \begin{minipage}[tb]{0.32\textwidth}
    \includegraphics[width=1\textwidth]{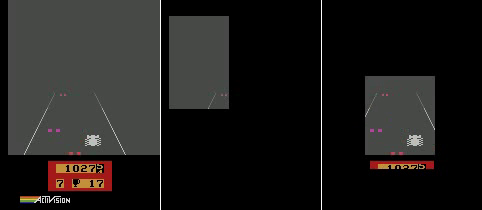}
    \end{minipage}
    \label{fig:en-9}
  }
  \caption{Visualizing the game Enduro. These examples are meant to be illustrative of the agent's strategies, which are used at different instances of time during the game. (a)-(c) illustrate the general strategy. (d)-(f) depict a special game stage when it is near the end of the day. (g), (h) and (i) also illustrate special game stages, which we name as `slacking,' `prepping,' and `smog,' respectively. More details are in the text below.}
  \label{fig:enduro}
\end{figure*}
The Atari 2600 game environments are provided by the Arcade Learning Environment~\cite{atari-py}, which is designed to be a rigorous evaluation system for general-purpose artificial intelligence (AI) algorithms.
The history of Atari games dates back to the 70s, when the game code was distributed in cartridges and played on game consoles with only 128 bytes of memory.
Since then, numerous variations of the original games like Ms. Pac-Man and Space Invaders have been created, which remain popular today.
While these games are designed to be challenging for human players, they only require a joystick and a few buttons to play.
The simple interface and challenging gameplay make these games suitable test beds for evaluating the learning, modeling, and planning capabilities of an AI agent.
With the rise of DeepRL, the Atari games have been used extensively for evaluating the newest DeepRL agents.
In this section, we select three popular games from Atari 2600 and illustrate the interpretation power of RS-Rainbow on these games via detailed case studies.

\subsection{Enduro} \label{sec:enduro-analysis}
Enduro is a racing game in which the total return for the player is the total number of cars that the player is able to pass while they control a vehicle.
The duration of the game is divided into `days.'
On each day, the player has to pass a minimum number of cars to qualify for the race next day.
If at any time, a collision with other cars happen, the player's vehicle suffers a severe speed penalty.
Passing more cars than the minimum required amount for each day does not bring extra return.
Variations in weather conditions and the time of the day add more challenges to the game.

By setting the number of score maps ($N$) to $2$ in RS-Rainbow, we can obtain two individual \emph{gazes} of the agent.
A gaze is a region assigned the highest importance and contributes the most to the Q value estimation.
We first describe the most common patterns followed by the two gazes throughout the game and use them to characterize the general policy.
Then we turn to special cases where gazes detect new patterns, which we discover to be associated to interesting deviations from the general policy.

\vspace{1ex}\noindent\textbf{General strategy}.
As shown in Figures~\ref{fig:en-1} to~\ref{fig:en-3}, both the left and the right gazes attend to the race track, but focus on different things.
We discover that at different instances of time, the left gaze focuses on parts of the race track that are at different distances, \emph{i.e.}, the far, the intermediate, and the closest sections.
In the meanwhile, the right gaze is locked on the player-controlled vehicle and follows it all the time.
For this reason, we refer to the right gaze as the player tracker.
Importantly, the locations that the left gaze is interested in nearly always contain cars, which are targets that should not be run into; and the player tracker (the right gaze), in addition to following the player's vehicle, monitors the closest upcoming cars, which may pose imminent collision threats.

The general rationale of RS-Rainbow when it is making a decision is summarized as follows.
On the one hand, the agent is tracking the player vehicle at all times, never losing focus on it, which means that the position of the player-controlled car and its surroundings are a major contributing factor to the decision made at each time step.
On the other hand, the agent is closely observing the conditions of the road, \emph{i.e.}, upcoming cars far or close, to successfully avoid collisions, which is actually the key to achieving a high score in this game.

We want to highlight three aspects in our interpretations.
First, the gazes are automatically learned without extra supervision.
Second, our interpretations are not a-posteriori analyses as is the case in~\cite{icml18} and~\cite{icml16}.
Instead, we outline attention patterns that provide information about what may contribute the most to decision making.
Third, the model's capability of producing informative visualizations leads to performance improvements, as we show via experiments in Section~\ref{sec:experiments}.

\vspace{1ex}\noindent\textbf{Counting down}.~At the bottom of the screen, there is a number that indicates the number of cars left to pass for achieving victory at the current level (this number is $13$, $12$, and $11$ in Figures~\ref{fig:en-4} to~\ref{fig:en-6}, respectively).
Once this number reaches $0$, a set of flags indicating victorious status would appear in place of the number, at which time the agent does not need to play the game anymore and can simply wait for a new day to begin.

We observe a new distribution of the left gaze when it is near the completion of the current level, which is different from what was found for the general strategy.
As shown in Figures~\ref{fig:en-4} to~\ref{fig:en-6}, the left gaze loses its focus on cars and diverts to the mileage and record board starting when there are $13$ cars left to pass before it can `rest.'
We say that the agent is `counting down' toward victory and `celebrating' in advance, like a runner before the finish line.
At the same time, the right gaze is still a player tracker, focusing on the player-controlled car.
This diversion of attention does not seem to impact the performance of the agent, though, as we never observed collisions at this stage, while the agent was also always able to pass the current level if it got here (note that there is a chance that this is due to environment determinism).
In any case, this pattern is internal, that the agent takes notice of the nearness of victory, and it would not have been revealed without the proposed region-sensitive module.

\vspace{1ex}\noindent\textbf{Slacking}.~Upon completing the current level, the agent won't receive any rewards for keep playing the game, until the next day starts.
During this period, the agent learns to output no-op actions, equaling not playing the game.
We refer to this behavior as ``slacking."
We are interested in what leads to the decision of not playing, or what the agent is looking at when not playing.
Figure~\ref{fig:en-7} can offer us some clues.
It shows that when the agent slacks off, both gazes fixate on the mileage board, where there are flags indicating task completion.
This means that the agent no longer considers the race track relevant, and instead finds the flags important for making a decision.
Eventually, the recognition that the flags are a sign of zero return leads to the no-op policy.

\begin{figure*}[tb]
  \centering
  \subfloat[Detecting Ghosts]{
    \begin{minipage}[tb]{0.32\textwidth}
    \includegraphics[width=1\textwidth]{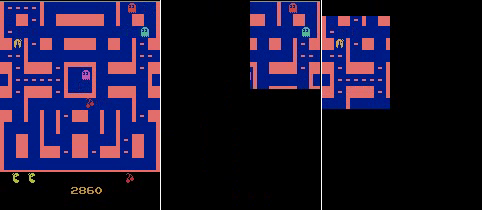}
    \end{minipage}
    \label{fig:ghost}
  }
  \subfloat[Detecting Vulnerable Ghosts]{
    \begin{minipage}[tb]{0.32\textwidth}
    \includegraphics[width=1\textwidth]{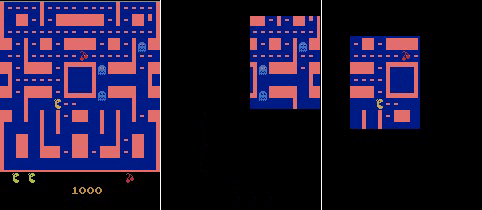}
    \end{minipage}
    \label{fig:vulnerable}
  }
  \subfloat[Detecting Fruits]{
    \begin{minipage}[tb]{0.32\textwidth}
    \includegraphics[width=1\textwidth]{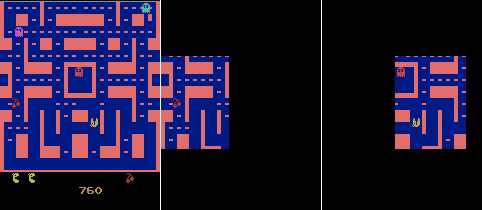}
    \end{minipage}
    \label{fig:fruit}
  }
  
  \subfloat[The Warp Tunnel]{
    \begin{minipage}[tb]{0.32\textwidth}
    \includegraphics[width=1\textwidth]{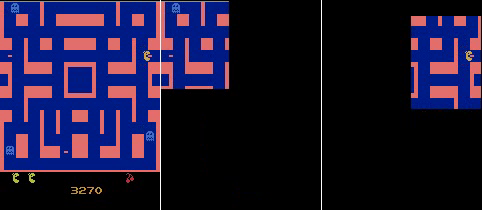}
    \end{minipage}
    \label{fig:warp-1}
  }
  \subfloat[The Warp Tunnel]{
    \begin{minipage}[tb]{0.32\textwidth}
    \includegraphics[width=1\textwidth]{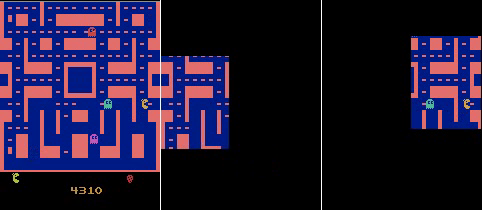}
    \end{minipage}
    \label{fig:warp-2}
  }
  \subfloat[The Warp Tunnel]{
    \begin{minipage}[tb]{0.32\textwidth}
    \includegraphics[width=1\textwidth]{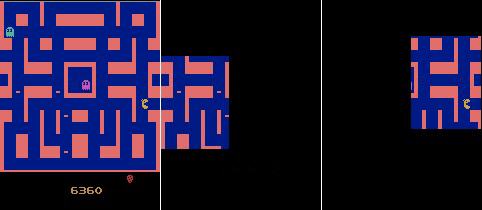}
    \end{minipage}
    \label{fig:warp-3}
  }
  
  \subfloat[The Last Pellet]{
    \begin{minipage}[tb]{0.32\textwidth}
    \includegraphics[width=1\textwidth]{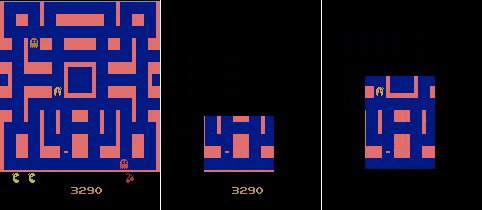}
    \end{minipage}
    \label{fig:pellet-start}
  }
  \subfloat[The Last Pellet]{
    \begin{minipage}[tb]{0.32\textwidth}
    \includegraphics[width=1\textwidth]{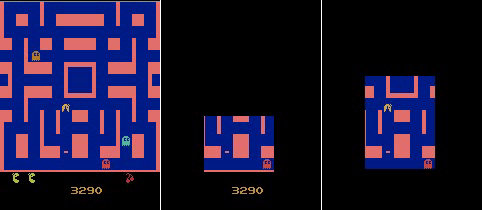}
    \end{minipage}
    \label{fig:pellet-ghost}
  }
  \subfloat[The Last Pellet]{
    \begin{minipage}[tb]{0.32\textwidth}
    \includegraphics[width=1\textwidth]{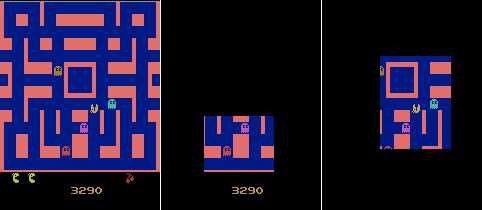}
    \end{minipage}
    \label{fig:pellet-end}
  }
  \caption{Visualizing the game Ms. Pac-Man. These examples are meant to be illustrative of the agent's strategies, which are used at different instances of time during the game. (a)-(c) Detecting moving objects: ghosts, vulnerable ghosts, and fruits. (d)-(f) Traveling through a warp tunnel. (g)-(i) Eating the last pellet in the maze.}
  \label{fig:pacman}
\end{figure*}
\vspace{1ex}\noindent\textbf{Prepping}.~Near the start of a new day (and new race, hence), the agent stops the previous slacking behavior, and accelerates the vehicle to a high speed in advance.
This gives it a head start when the other cars start to appear.
At this point, the flags are still up and there are still no rewards for playing.
Therefore, it is intriguing why the agent has changed its behavior.
The left gaze shown in Figure~\ref{fig:en-8} can offer some hints.
The left gaze is focused on an inconspicuous region in the background, containing some mountains and the sky.
As it turns out, the agent recognizes that it is dawn time (time near the start of a new race) from the unique appearances of the (light gray) sky and the (orange) mountains.
Since dawn indicates forthcoming rewards, the previously unimportant mountains and the sky become important at this time for value prediction.

\vspace{1ex}\noindent\textbf{Smog}.~During some periods of the game, the visibility of cars on the road becomes much worse---cars too far ahead become invisible, while those closer to the player-controlled vehicle have only their tail lights visible.
This resembles a foggy day when fog/smoke lowers visibility on the road.
We have observed minor performance decrease of the agent during these periods---it occasionally bumps into other cars and suffers a speed penalty.
As shown in Figure~\ref{fig:en-9}, when the smog obstructs the view of the agent, unlike in normal cases, its left gaze loses track of the cars ahead, and instead wanders off to empty fields outside the race track.
Therefore, it is possible for us to explain why the agent does not perform as well and have collisions with other cars---the analysis of its left gaze reveals that it has difficulty to localize the cars ahead, thus it is less efficient at preventing collisions.

In summary, we have derived an interpretation of the general strategy of the agent when playing the game Enduro, which is to track the player-controlled car, while at the same time, tracking the cars ahead that may cause collision.
We have also derived explanations for some interesting behaviors of the agent in certain special stages of the game, and we find most of those findings intuitive.

\begin{figure*}[tb]
  \centering
  \subfloat[Jumping]{
    \begin{minipage}[tb]{0.32\textwidth}
    \includegraphics[width=1\textwidth]{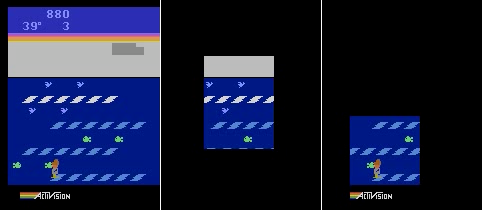}
    \end{minipage}
    \label{fig:jumping}
  }
  \subfloat[Inspecting Progress]{
    \begin{minipage}[tb]{0.32\textwidth}
    \includegraphics[width=1\textwidth]{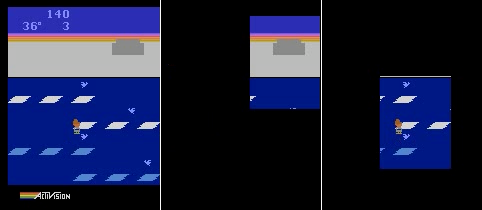}
    \end{minipage}
    \label{fig:inspecting}
  }
  \subfloat[Entering the Igloo]{
    \begin{minipage}[tb]{0.32\textwidth}
    \includegraphics[width=1\textwidth]{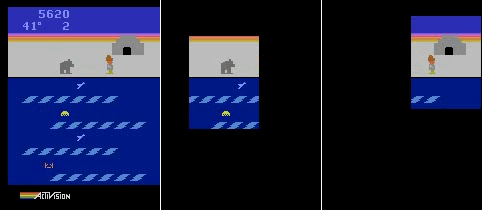}
    \end{minipage}
    \label{fig:entering}
  }
  \caption{Visualizing the game Frostbite. Each sub-figure depicts the view of the agent when it is controlling the character to (a) jump over ice blocks, (b) check the construction progress of the igloo, or (c) enter the igloo.}
  \label{fig:frostbite}
\end{figure*}

\subsection{Ms. Pac-Man}
In the game Ms. Pac-Man, the objective of the player is to control the character, Ms. Pac-Man, to accumulate points by collecting pellets in a maze, while at the same time avoiding ghosts.
Eating power pellets turns the ghosts into vulnerable ghosts that can be eaten for extra points.
Eating fruits also provides bonus points.
Therefore, recognizing the moving objects, \emph{i.e.}, the ghosts, the vulnerable ghosts, and the fruits, is essential for earning a high return.
Ms. Pac-Man proceeds to the next level after all pellets have been eaten.

Figure~\ref{fig:pacman} illustrates the learned gazes of RS-Rainbow in this game.
We find that during the entire game, the right gaze stays focused on Ms. Pac-Man to track its position and detect nearby threats and pellets (the right gaze in the game Enduro has a similar function), while the left gaze attends to different moving objects and locations in different game states.
Next, we provide interpretations of several game strategies employed by the agent via visualizations.

\vspace{1ex}\noindent\textbf{Moving object detection}.~As shown in Figure~\ref{fig:ghost}, the left gaze detects two ghosts in the upper-right corner of the maze.
At the same time, Ms. Pac-Man, located by the right gaze, travels in the mid-left region to collect rewards without any imminent threats.
At another instance, depicted by Figure~\ref{fig:vulnerable}, when Ms. Pac-Man chases after vulnerable ghosts, its left gaze locks in on its prey: three vulnerable ghosts in the mid-right region.
Similarly, in Figure~\ref{fig:fruit}, the left gaze detects a cherry that has just appeared near the lower-left warp tunnel entrance.
In the steps that follow (not depicted), Ms. Pac-Man enters the closest warp tunnel on the right to be transported to the cherry.

\vspace{1ex}\noindent\textbf{Traveling through a warp tunnel}.~Frequently in the game, the agent controls Ms. Pac-Man to travel through a warp tunnel in order to collect rewards on the other side of the maze.
During these times, the left gaze stays focused on the tunnel exit.
Figures~\ref{fig:warp-1},~\ref{fig:warp-2}, and~\ref{fig:warp-3} depict scenarios like that.
For instance, in Figure~\ref{fig:warp-2}, as the right gaze stays on Ms. Pac-Man, which is entering the middle tunnel from right, the left gaze focuses on the tunnel exit on the left.
This pattern suggests that the agent has planning in its `mind,'---it is aware where Ms. Pac-Man will be after warp tunnel transportation and looks at that location in advance.

\vspace{1ex}\noindent\textbf{Eating the last pellet}.~In Figures~\ref{fig:pellet-start},~\ref{fig:pellet-ghost}, and~\ref{fig:pellet-end}, we show that the left gaze focuses on the last pellet in the maze (near the bottom of the maze) during the entire time when Ms Pac-Man tries to collect that pellet.
In Figure~\ref{fig:pellet-start}, the left gaze locates the pellet for the first time and Ms. Pac-Man starts to move toward it.
A few steps later, as shown in Figure~\ref{fig:pellet-ghost}, a red ghost appears around the pellet, and this is detected by the left gaze.
Consequently, Ms. Pac-Man diverts to right to avoid the ghost.
Shortly after that, as depicted by Figure~\ref{fig:pellet-end}, several ghosts besiege Ms. Pac-Man, having approached it from different directions.
At this time the pellet is still not eaten and the left gaze stays on it (the pellet is blocked from view by the red ghost), but there is nothing left that Ms. Pac-Man can do.

\subsection{Frostbite}
In the game Frostbite, the player must control a character to build an igloo, and after building the igloo, to enter it.
These must be done before the temperature drops to zero.
The way to build the igloo is to jump on uncollected (white) floating ice blocks.
Collecting ice blocks, having a lot of temperature left, and catching fish all provide rewards.
Conversely, falling into water, temperature dropping to zero, and making contact with the bear, clams, or birds all cost the player a life.

Similarly, for this game, we find a division of responsibility between the two gazes.
While the right gaze always follows the player's character, the left gaze takes on the role of a generic object detector, which detects the most important object to the sub-task that is performed during a specific period.
In the following, we identify three types of such targets, each of which is key to achieving a sub-goal in the game, and we explain what the agent's strategy is in each situation, using visualizations shown in Figure~\ref{fig:frostbite}.

\vspace{1ex}\noindent\textbf{Jumping (the general strategy)}.~In this game, the most important skill to learn is to jump onto white ice blocks (which are moving).
Doing this provides immediate rewards as well as facilitates igloo building, which is a condition for advancing to the next level.
As a result, the type of left gaze that we observe in Figure~\ref{fig:jumping} is the most common one throughout the game---it detects white ice blocks that are the next jumping destinations.
We observe that for most of the time during the game, the left gaze locks on white ice blocks, as the right gaze tracks the character.
And this pattern accompanies the agent's policy of jumping around and collecting new ice blocks.
Hence, we summarize the agent's main strategy and the rationale as making jumps based on the locations of uncollected ice blocks and the character.

\vspace{1ex}\noindent\textbf{Inspecting progress}.~Quite curiously, another pattern of the left gaze that we observe during the game is that it regularly lands upon the igloo in construction (Figure~\ref{fig:inspecting}), located at the top-right corner of the screen.
This can be interpreted as the agent checking up on the construction of the igloo.
This behavior becomes especially prominent as the igloo comes close to being finished.
Note that once the igloo is fully built, the character can enter it to advance to the next level.
During the entire time, the right gaze focuses on the character as usual.
We think that in this case, it is intuitive to interpret from what we observe that the agent `understands' the logic of the game, especially the importance of the igloo for game continuation, and that it indeed relies on observing the status of the igloo for advancing to the next level.

\vspace{1ex}\noindent\textbf{Entering the igloo}.~After the igloo is built, the character needs to enter it while avoiding the bear.
Figure~\ref{fig:entering} depicts a moment when the agent controls the character to run to the igloo.
During the entire process of jumping onto land and running to the igloo, the left gaze follows the bear, while the right gaze tracks the character.
The left and the right gazes suggest that the strategy of the agent at this stage of the game is to monitor the position of the bear as well as the position of the character, while controlling the character to run to the igloo.

\section{Quantitative Evaluation} \label{sec:experiments}
\begin{table*}[t]
\small
\caption{Comparison with state-of-the-art methods under the no-op condition. $^*$ denotes our re-implementation.}
\label{table:1}
\centering
\begin{tabular}{p{2.5cm}cccccccccr}
\toprule
Method & beam\_rider & breakout & enduro & frostbite & ms\_pacman & pong & seaquest & space\_invaders\\ 
\midrule
DQN & 8,627.5 & 385.5 & 729.0 & 797.4 & 3,085.6 & 19.5 & 5,860.6 & 1,692.3\\
DDQN & 13,772.8 & 418.5 & 1,211.8 & 1,683.3 & 2,711.4 & 20.9 & 16,452.7 & 2,525.5\\
Prior. DDQN & 22,430.7 & 381.5 & 2,155.0 & 3,421.6 & 4,751.2 & 20.7 & 44,417.4 & 7,696.9\\
Duel. DDQN  & 12,164.0 & 345.3 & 2,258.2 & 4,672.8 & 6,283.5 &\textbf{21.0} & 50,254.2 & 6,427.3\\
Dist. DQN & 13,213.4 & \textbf{612.5} & 2,259.3 & 3,938.2 & 3,769.2 & 20.8 & 4,754.4 & 6,869.1\\
Noisy DQN & 12,534.0 & 459.1 & 1,129.2 & 583.6 & 2,501.6 & \textbf{21.0} & 2,495.4 & 2,145.5\\
Rainbow & 16,850.2 & 417.5 & 2,125.9 & 9,590.5 & 5,380.4 & 20.9 & 15,898.9 & 18,789.0\\
Rainbow* & 17,656.8 & 370.7 & 2,283.6 & 11,298.3 & 6,686.3 & 20.9 & 73,601.4 & 3,001.2\\
RS-Rainbow (\textbf{ours}) & \textbf{26,722.3} & 434.2 & \textbf{2,329.1} & \textbf{12,902.0} & \textbf{7,219.3} & 20.9 & \textbf{245,307.3} & \textbf{19,670.0}\\
\bottomrule
\end{tabular}
\end{table*}

\subsection{Evaluation Benchmarks and Pre-Processing}
A suite of $57$ Atari 2600 games from the Arcade Learning Environment~\cite{atari-py} are commonly used to benchmark DeepRL algorithms.
Due to limited computational resources, we conduct experiments on $8$ of the more popular games, and compare the proposed method to other state-of-the-art methods.

We follow the steps in~\cite{duel, per, rainbow} to pre-process the input frames, where each frame is transformed into single-channel grayscale and resized to the resolution of $84\times84$.
At each time step, the input is four consecutive frames stacked along the channel dimension.

\subsection{Implementation Details} \label{sec:implementation}
We use an open-source implementation of Rainbow with the same hyperparameters and model configuration as in the original work~\cite{rainbow}.
For the selection of normalization layers in the region-sensitive module, we employ the sigmoid function in games Breakout, Space Invaders, and Seaquest, and the softmax function in the rest of the games.
During both training and testing, we cap the episode length at $108$K frames and adopt an action repeat of $4$.
During training, rewards are clipped to the range of $[-1,1]$, and exploration is done by adopting the noisy linear layers.
For every $100K$ environment steps, we suspend training and evaluate the agent for $10$ episodes, and use the snapshot with the highest average score for testing.
At test time, we adopt the $\epsilon$-greedy policy with $\epsilon=0.001$.
We evaluate the agents under the no-op random start condition, therefore, at the beginning of each test episode, a random number (up to $30$) of no-ops are executed (as is done in training) before the agent starts playing.
To report the final results on each game, we initialize the environment three times with three random seeds, and with each seed, run $200$ test episodes; in Table~\ref{table:1}, each reported score is obtained by first averaging scores across $200$ test episodes for each seed and then averaging again across the three random seeds.

\subsection{Comparison with State-of-the-Art Methods}
In Table~\ref{table:1}, we evaluate the performance of RS-Rainbow against several state-of-the-art DQN methods.
These include Rainbow~\cite{rainbow}, distributional DQN~\cite{distributional}, noisy DQN~\cite{noisy}, duelling DDQN~\cite{duel}, prioritized DDQN~\cite{per}, DDQN~\cite{double-dqn}, and DQN~\cite{dqn15}.
For the performance of Rainbow, we report both the original scores quoted from~\cite{rainbow} and the ones reproduced by us.
We denote our implementation as Rainbow$^*$ in Table~\ref{table:1}.
Our results of RS-Rainbow and Rainbow$^*$ are obtained after training the models for $50$ million steps on each game, with an exception to Frostbite, for which we train the models for $100$ million steps.

As shown in Table~\ref{table:1}, RS-Rainbow outperforms Rainbow$^*$ with a large margin in $7$ out of $8$ games.
Specifically, relative to Rainbow$^*$, RS-Rainbow achieves a $51\%$ improvement in the game Beam Rider, $17\%$ improvement in Breakout, $2\%$ improvement in Enduro, $14\%$ improvement in Frostbite, $8\%$ improvement in Ms. Pac-Man, $233\%$ improvement in Seaquest, and $555\%$ improvement in Space Invaders.
In the game Pong, RS-Rainbow obtains the same score as Rainbow$^*$, which is a nearly perfect $20.9$, where $21$ is the highest achievable score.

Compared with other methods, RS-Rainbow outperforms the respective best-performing models in $6$ out of $8$ games by solid margins, including Beam Rider, Enduro, Frostbite, Ms. Pac-Man, Seaquest, and Space Invaders.
For instance, it leads prioritized DDQN in Beam Rider by $19\%$, duelling DDQN in Ms. Pac-Man and Seaquest by $15\%$ and $388\%$, respectively, and Rainbow in Frostbite and Space Invaders by $35\%$ and $5\%$, respectively.
On the rest two games,~\emph{i.e.}, Breakout and Pong, RS-Rainbow also obtains competitive scores.

\subsection{Extension to Gym Retro}
We extend the quantitative evaluation of RS-Rainbow to the domain of Gym Retro~\cite{nichol2018retro}, a platform which allows the conversion of ROMs of proprietary classic video games into Gym environments for academic purposes.
We choose the game Airstriker-Genesis for evaluation, which has a non-commercial ROM available while at the same time is a challenging and popular game.
The experiment configuration remains the same as in Atari games, except that we train RS-Rainbow for just $20$ million steps instead of $50$ million, at which point it already outperforms Rainbow.
As shown in Table~\ref{table:2}, the proposed RS-Rainbow outperforms Rainbow in the game Airstriker-Genesis by $25\%$ (relative), showing consistency with our observation in Table~\ref{table:1}.
\begin{table}[t]
\small
\caption{Comparison between RS-Rainbow and Rainbow on the game Airstriker-Genesis from Gym Retro.}
\label{table:2}
\centering
\begin{tabular}{p{3.5cm}cr}
\toprule
Method & Airstriker-Genesis\\ 
\midrule
Rainbow & 3,068.6\\
RS-Rainbow (\textbf{ours}) & \textbf{3,850.4}\\
\bottomrule
\end{tabular}
\end{table}

\section{Conclusion and Future Work} \label{sec:conclusion}
In this work, we have approached the problem of interpreting DeepRL models from a learning perspective.
Instead of visualizing learned policies a-posteriori as contemporary work does, we propose RS-Rainbow, a simple architecture with embedded interpretability.
As we demonstrate in a series of visual analyses performed on various games, by backpropagating gradients from the proposed region-sensitive module to the input frames, we can easily identify regions that are the most important to decision making.
Moreover, the proposed RS-Rainbow achieves state-of-the-art performance on several games from the Atari 2600 benchmark suite~\cite{atari-py}.

More specifically, by conducting case studies on three representative Atari games, we show how the region-sensitive module is able to provide hints on both the general strategies as well as specialized strategies that are employed by the agent.
As patterns in these visualizations can be easily interpreted by humans, they lend credibility to the outputs of the black-box agent, which is critical to applications where obtaining human trust is an important consideration, or even prerequisite, for real-world deployment.
In addition, the proposed region-sensitive module may provide information about why the agent underperforms in certain situations.
For instance, due to a failure to detect key objects in certain stages of a game.

There can potentially be many enhancements to the current model that would be interesting and useful.
We notice that the current visualization spans a large area and cannot precisely pinpoint individual objects.
This might be addressed by employing multi-scale feature maps, or incorporating dilated convolutions~\cite{deeplabv2}, which prove to be effective at maintaining spatial details while capturing large-scale information.
Then with a multi-scale, high-resolution visualization, we may be able to conduct object-level (instead of region-level) analyses, reaching toward more accurate and insightful interpretations.

As we design the region-sensitive module to be general to CNN-based models, it will be interesting to investigate the effectiveness of our proposal in other state-of-the-art DeepRL methods, such as asynchronous advantage actor-critic~\cite{a3c16}, proximal policy optimization~\cite{ppo}, and the deep deterministic policy gradient (DDPG) algorithm~\cite{ddpg}.
These methods belong to the paradigm of policy optimization~\cite{sutton}, which aims to arrive at the optimal policy by performing gradient ascent in a policy network, and are fundamentally different from the value-based DQN algorithm.
In addition, DDPG is designed for addressing the control problem in a continuous domain, where actions are represented by real-valued random variables instead of discrete ones.
For instance, driving a car from a first-person view requires the model to predict real-valued variables such as the turning angle of the steering wheel and the magnitude of forces to apply on the brake and the gas pedals.

Moreover, in the past year, we have witnessed the emergence of several large-scale deep multi-agent reinforcement learning algorithms, which are proposed to address more complex tasks that involve the competition and cooperation between multiple intelligent agents in an environment.
For instance, the multi-agent algorithm OpenAI Five~\cite{openaifive} trains five independent advantage actor-critic agents, which learn to work as a team to play the game Dota 2 and have defeated human champions.
Some other examples of deep multi-agent algorithms that achieve human-level performance in complex games are AlphaStar~\cite{alphastar} which plays StarCraft II, FTW~\cite{ftw} which plays the 3D multi-player first-person video game, Quake III Arena, and Pluribus~\cite{texasholdem} which plays multi-player poker and have defeated human professionals in no-limit Texas hold’em.
In principle, our proposed region-sensitive module works with any agent that accepts visual inputs, therefore, it would be interesting to apply it to some of these deep multi-agent reinforcement learning algorithms, and observe if visualizations of individual agents lead to insights on multi-player dynamics.

Another recent development in DeepRL is the exploitation of massively distributed CPUs for highly paralleled exploration and the decoupled learning of `actors' and a `learner,' such as in Ape-X DQN~\cite{apex}.
Since the highest scores in the Atari 2600 games are achieved by Ape-X DQN, it would be necessary to test the effectiveness of the proposed region-sensitive module in this model.
However, the scale of parallelization in Ape-X DQN is achieved with hundreds of CPU cores, which are beyond our computational capacity.

We leave these explorations to future opportunities.

\vspace{1ex}\noindent\textbf{Acknowledgements}.
This work was supported by the ERC grant ERC-2012-AdG 321162-HELIOS, EPSRC grant Seebibyte EP/M013774/1, EPSRC/MURI grant EP/N019474/1, and Tencent.
We would also like to acknowledge the Royal Academy of Engineering and Five AI.

\bibliographystyle{IEEEtran}
\bibliography{main}

\end{document}